%% file: main.tex
\documentclass{article}

\usepackage[preprint, nonatbib]{nips_2018}

\usepackage[utf8]{inputenc} 
\usepackage[T1]{fontenc}    
\usepackage{url}            
\usepackage{booktabs}       
\usepackage{amsfonts}       
\usepackage{nicefrac}       
\usepackage{microtype}      
\usepackage{color}
\usepackage[]{graphicx}
\usepackage{amsmath}
\usepackage{xcolor}
\usepackage{paralist}
\usepackage{enumerate}
\usepackage[]{subcaption}
\usepackage{siunitx}
\usepackage{makecell}
\usepackage{threeparttable}
\usepackage{tikz}
\usepackage{multirow}
\usepackage{tabularx} 
\usepackage{soul}

\title{Representation Learning with \\Contrastive Predictive Coding}

\author{
  Aaron van den Oord \\
  DeepMind \\
  \texttt{avdnoord@google.com} \\
  \And
  Yazhe Li \\
  DeepMind \\
  \texttt{yazhe@google.com} \\
  \And
  Oriol Vinyals \\
  DeepMind \\
  \texttt{vinyals@google.com} \\
}

\begin{document}

\maketitle

\begin{abstract}

While supervised learning has enabled great progress in many applications, unsupervised learning has not seen such widespread adoption, and remains an important and challenging endeavor for artificial intelligence.
In this work, we propose a universal unsupervised learning approach to extract useful representations from high-dimensional data, which we call Contrastive Predictive Coding.
The key insight of our model is to learn such representations by predicting the future in \emph{latent} space by using powerful autoregressive models. We use a probabilistic contrastive loss which induces the latent space to capture information that is maximally useful to predict future samples. It also makes the model tractable by using negative sampling.
While most prior work has focused on evaluating representations for a particular modality, we demonstrate that our approach is able to learn useful representations achieving strong performance on four distinct domains: speech, images, text and reinforcement learning in 3D environments.
\end{abstract}

\input{introduction}

\input{method}
\input{experiments}
\input{conclusion}

\bibliographystyle{unsrt}
\bibliography{main}

\input{appendix}

\end{document}

%% file: introduction.tex
\section{Introduction}

Learning high-level representations from labeled data with layered differentiable models in an end-to-end fashion is one of the biggest successes in artificial intelligence so far. These techniques made manually specified features largely redundant and have greatly improved state-of-the-art in several real-world applications \cite{krizhevsky2012imagenet, hinton2012deep, sutskever2014sequence}. However, many challenges remain, such as data efficiency, robustness or generalization. 

Improving representation learning requires features that are less specialized towards solving a single supervised task. For example, when pre-training a model to do image classification, the induced features transfer reasonably well to other image classification domains, but also lack certain information such as color or the ability to count that are irrelevant for classification but relevant for e.g. image captioning \cite{showandtell}. Similarly, features that are useful to transcribe human speech may be less suited for speaker identification, or music genre prediction. Thus, unsupervised learning is an important stepping stone towards robust and generic representation learning.

Despite its importance, unsupervised learning is yet to see a breakthrough similar to supervised learning: modeling high-level representations from raw observations remains elusive. Further, it is not always clear what the ideal representation is and if it is possible that one can learn such a representation without additional supervision or specialization to a particular data modality.

One of the most common strategies for unsupervised learning has been to predict future, missing or contextual information. This idea of predictive coding \cite{elias1955predictive, atal1970adaptive} is one of the oldest techniques in signal processing for data compression. In neuroscience, predictive coding theories suggest that the brain predicts observations at various levels of abstraction \cite{rao1999predictive, friston2005theory}. Recent work in unsupervised learning has successfully used these ideas to learn word representations by predicting neighboring words \cite{mikolov2013efficient}. For images, predicting color from grey-scale or the relative position of image patches has also been shown useful \cite{zhang2016colorful, Doersch_2015_ICCV}.
We hypothesize that these approaches are fruitful partly because the context from which we predict related values are often conditionally dependent on the same shared high-level latent information. And by casting this as a prediction problem, we automatically infer these features of interest to representation learning.

In this paper we propose the following: first, we compress high-dimensional data into a much more compact latent embedding space in which conditional predictions are easier to model. Secondly, we use powerful autoregressive models in this latent space to make predictions many steps in the future. Finally, we rely on Noise-Contrastive Estimation \cite{gutmann2010noise} for the loss function in similar ways that have been used for learning word embeddings in natural language models, allowing for the whole model to be trained end-to-end. We apply the resulting model, Contrastive Predictive Coding (CPC) to widely different data modalities, images, speech, natural language and reinforcement learning, and show that the same mechanism learns interesting high-level information on each of these domains, outperforming other approaches.

\begin{figure*}[t!]
  \center
  \includegraphics[width=0.90\textwidth]{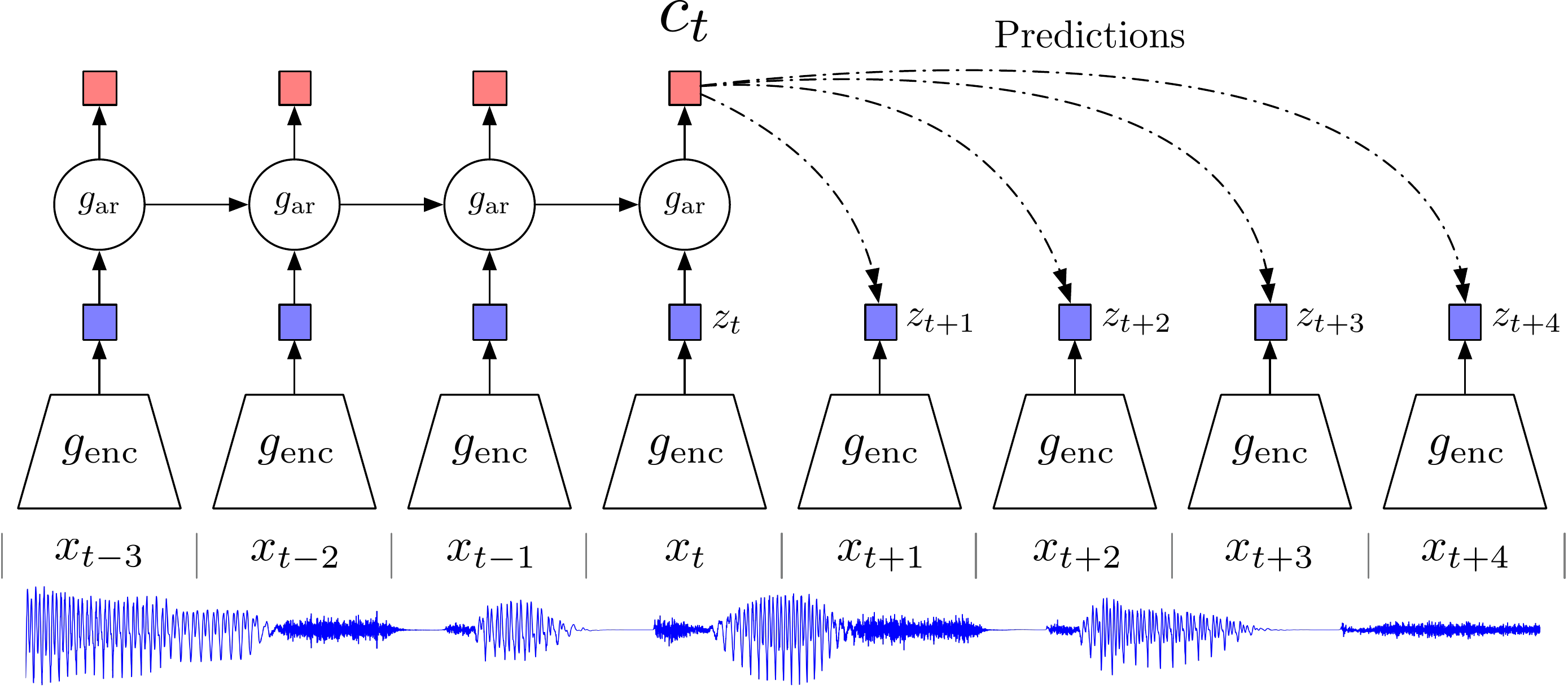}
  \caption{Overview of Contrastive Predictive Coding, the proposed representation learning approach. Although this figure shows audio as input, we use the same setup for images, text and reinforcement learning.}
  \label{fig:overview}
\end{figure*}

%% file: method.tex
\section{Contrastive Predicting Coding}
\label{method}

We start this section by motivating and giving intuitions behind our approach. Next, we introduce the architecture of Contrastive Predictive Coding (CPC). After that we explain the loss function that is based on Noise-Contrastive Estimation. Lastly, we discuss related work to CPC.

\subsection{Motivation and Intuitions}

The main intuition behind our model is to learn the representations that encode the underlying shared information between different parts of the (high-dimensional) signal. At the same time it discards low-level information and noise that is more local.
In time series and high-dimensional modeling, approaches that use next step prediction exploit the local smoothness of the signal.
When predicting further in the future, the amount of shared information becomes much lower, and the model needs to infer more global structure. These 'slow features' \cite{wiskott2002slow} that span many time steps are often more interesting (e.g., phonemes and intonation in speech, objects in images, or the story line in books.).

One of the challenges of predicting high-dimensional data is that unimodal losses such as mean-squared error and cross-entropy are not very useful, and powerful conditional generative models which need to reconstruct every detail in the data are usually required. But these models are computationally intense, and waste capacity at modeling the complex relationships in the data $x$, often ignoring the context $c$. For example, images may contain thousands of bits of information while the high-level latent variables such as the class label contain much less information (10 bits for 1,024 categories). This suggests that modeling $p(x|c)$ directly may not be optimal for the purpose of extracting shared information between $x$ and $c$. When predicting future information we instead encode the target $x$ (future) and context $c$ (present) into a compact distributed vector representations (via non-linear learned mappings) in a way that maximally preserves the mutual information of the original signals $x$ and $c$ defined as 
\begin{align}
I(x;c) &= \sum_{x,c} p(x,c) \log \frac{p(x|c)}{p(x)}. \label{mi}
\end{align}
By maximizing the mutual information between the encoded representations (which is bounded by the MI between the input signals), we extract the underlying latent variables the inputs have in commmon. 

\subsection{Contrastive Predictive Coding}
\label{model_math}

Figure \ref{fig:overview} shows the architecture of Contrastive Predictive Coding models. First, a non-linear encoder $g_{\text{enc}}$ maps the input sequence of observations $x_t$ to a sequence of latent representations $z_t=g_{\text{enc}}(x_t)$, potentially with a lower temporal resolution. Next, an autoregressive model $g_{\text{ar}}$ summarizes all $z_{\leq t}$ in the latent space and produces a context latent representation $c_t=g_{\text{ar}}(z_{\leq t})$.

As argued in the previous section we do not predict future observations $x_{t+k}$ directly with a generative model $p_k(x_{t+k}| c_t)$. Instead we model a density ratio which preserves the mutual information between $x_{t+k}$ and $c_t$ (Equation~\ref{mi}) as follows (see next sub-section for further details):
\begin{equation}
f_k(x_{t+k}, c_{t}) \propto \frac{p(x_{t+k}| c_t)}{p(x_{t+k})} \label{math:density_ratio}
\end{equation}
where $\propto$ stands for 'proportional to' (i.e. up to a multiplicative constant). Note that the density ratio $f$ can be unnormalized (does not have to integrate to 1). Although any positive real score can be used here, we use a simple log-bilinear model:
\begin{align}
f_k(x_{t+k}, c_{t}) = \exp \Big(z_{t+k}^T W_k c_t\Big),
\end{align}
In our experiments a linear transformation $W_k^T c_t$ is used for the prediction with a different $W_k$ for every step $k$. Alternatively, non-linear networks or recurrent neural networks could be used.

By using a density ratio $f(x_{t+k}, c_{t})$ and inferring $z_{t+k}$ with an encoder, we relieve the model from modeling the high dimensional distribution $x_{t_k}$. Although we cannot evaluate $p(x)$ or $p(x|c)$ directly, we can use samples from these distributions, allowing us to use techniques such as Noise-Contrastive Estimation \cite{gutmann2010noise, mnih2012fast, jozefowicz2016exploring} and Importance Sampling \cite{bengio2008is} that are based on comparing the target value with randomly sampled negative values.

In the proposed model, either of $z_t$ and $c_t$ could be used as representation for downstream tasks.
The autoregressive model output $c_t$ can be used if extra context from the past is useful. One such example is speech recognition, where the receptive field of $z_t$ might not contain enough information to capture phonetic content. In other cases, where no additional context is required, $z_t$ might instead be better. If the downstream task requires one representation for the whole sequence, as in e.g. image classification, one can pool the representations from either $z_t$ or $c_t$ over all locations. 

Finally, note that any type of encoder and autoregressive model can be used in the proposed framework. For simplicity we opted for standard architectures such as strided convolutional layers with resnet blocks for the encoder, and GRUs \cite{cho2014learning} for the autoregresssive model. More recent advancements in autoregressive modeling such as masked convolutional architectures \cite{oord2016wavenet, aaron2016pixelcnn} or self-attention networks \cite{attentionNIPS2017} could help improve results further.

\subsection{InfoNCE Loss and Mutual Information Estimation}
\label{sec:nce}

Both the encoder and autoregressive model are trained to jointly optimize a loss based on NCE, which we will call InfoNCE. Given a set $X=\{x_1, \dots x_{N}\}$ of $N$ random samples containing one positive sample from $p(x_{t+k}|c_t)$ and $N-1$ negative samples from the 'proposal' distribution $p(x_{t+k})$, we optimize:
\begin{align}
\mathcal{L_{\text{N}}} &= - \mathop{{}\mathbb{E}}_{X}\left[\log \frac{f_k(x_{t+k}, c_{t})}{\sum_{x_j \in X} f_k(x_j, c_{t})}\right] \label{loss}
\end{align}

Optimizing this loss will result in $f_k(x_{t+k}, c_{t})$ estimating the density ratio in equation \ref{math:density_ratio}. This can be shown as follows.

The loss in Equation \ref{loss} is the categorical cross-entropy of classifying the positive sample correctly, with $\frac{f_k}{\sum_{X} f_k}$ being the prediction of the model. Let us write the optimal probability for this loss as $p(d=i|X, c_{t})$ with $[d=i]$ being the indicator that sample $x_i$ is the 'positive' sample. The probability that sample $x_i$ was drawn from the conditional distribution $p(x_{t+k}|c_{t})$ rather than the proposal distribution $p(x_{t+k})$ can be derived as follows:
\begin{align}
p(d = i| X, c_{t}) &= \frac{p(x_i|c_{t})\prod_{l\neq i}p(x_l)}{\sum^N_{j=1} p(x_j|c_{t})\prod_{l\neq j}p(x_l)} \nonumber \\
&=\frac{\frac{p(x_i|c_{t})}{p(x_i)}}{\sum^N_{j=1} \frac{p(x_j|c_{t})}{p(x_j)}}.
\end{align}
As we can see, the optimal value for $f(x_{t+k}, c_{t})$ in Equation \ref{loss} is proportional to $\frac{p(x_{t+k}|c_{t})}{p(x_{t+k})}$ and this is independent of the the choice of the number of negative samples $N-1$.

Though not required for training, we can evaluate the mutual information between the variables $c_t$ and $x_{t+k}$ as follows:
$$
I(x_{t+k}, c_t) \geq \log(N)-\mathcal{L_{\text{N}}},
$$
which becomes tighter as N becomes larger. Also observe that minimizing the InfoNCE loss $\mathcal{L_{\text{N}}}$ maximizes a lower bound on mutual information. For more details see Appendix. 

\subsection{Related Work}
\label{related}

CPC is a new method that combines predicting future observations (predictive coding) with a probabilistic contrastive loss (Equation \ref{loss}). This allows us to extract slow features, which maximize the mutual information of observations over long time horizons. Contrastive losses and predictive coding have individually been used in different ways before, which we will now discuss.

Contrastive loss functions have been used by many authors in the past. For example, the techniques proposed by \cite{chopra2005learning, weinberger2009distance, schroff2015facenet} were based on triplet losses using a max-margin approach to separate positive from negative examples. More recent work includes Time Contrastive Networks \cite{sermanet2017time} which proposes to minimize distances between embeddings from multiple viewpoints of the same scene and whilst maximizing distances between embeddings extracted from different timesteps. In Time Contrastive Learning \cite{NIPS2016_6395} a contrastive loss is used to predict the segment-ID of multivariate time-series as a way to extract features and perform nonlinear ICA.

There has also been work and progress on defining prediction tasks from related observations as a way to extract useful representations, and many of these have been applied to language. In Word2Vec \cite{mikolov2013efficient} neighbouring words are predicted using a contrastive loss. Skip-thought vectors \cite{kiros2015skip} and Byte mLSTM \cite{radford2017learning} are alternatives which go beyond word prediction with a Recurrent Neural Network, and use maximum likelihood over sequences of observations. In Computer Vision \cite{wang2015unsupervised} use a triplet loss on tracked video patches so that patches from the same object at different timesteps are more similar to each other than to random patches. \cite{Doersch_2015_ICCV, noroozi2016unsupervised} propose to predict the relative postion of patches in an image and in \cite{zhang2016colorful} color values are predicted from a greyscale images.

%% file: experiments.tex
\section{Experiments}
\label{experiments}

We present benchmarks on four different application domains: speech, images, natural language and reinforcement learning. For every domain we train CPC models and probe what the representations contain with either a linear classification task or qualitative evaluations, and in reinforcement learning we measure how the auxiliary CPC loss speeds up learning of the agent.

\subsection{Audio}
\label{audio_exp}

\begin{figure}[t]

\begin{minipage}{0.47\textwidth}
  \centering
  \includegraphics[width=0.99\textwidth]{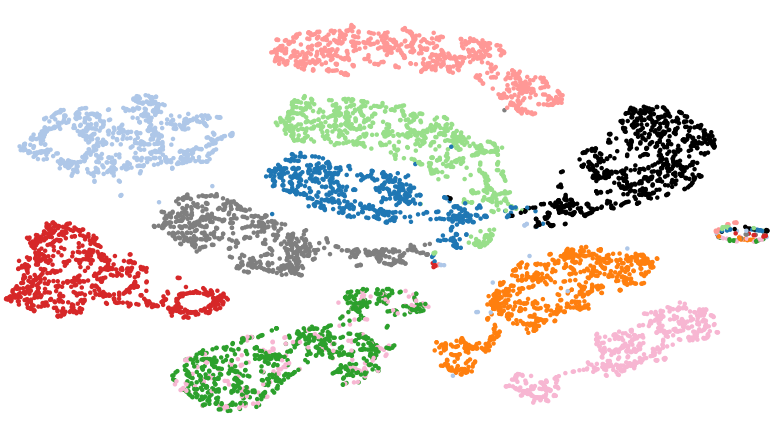}
  \caption{t-SNE visualization of audio (speech) representations for a subset of 10 speakers (out of 251). Every color represents a different speaker.}
  \label{speaker_tsne}
\end{minipage}%
\hfill
\begin{minipage}{0.47\textwidth}
  \centering
  \includegraphics[width=0.99\textwidth]{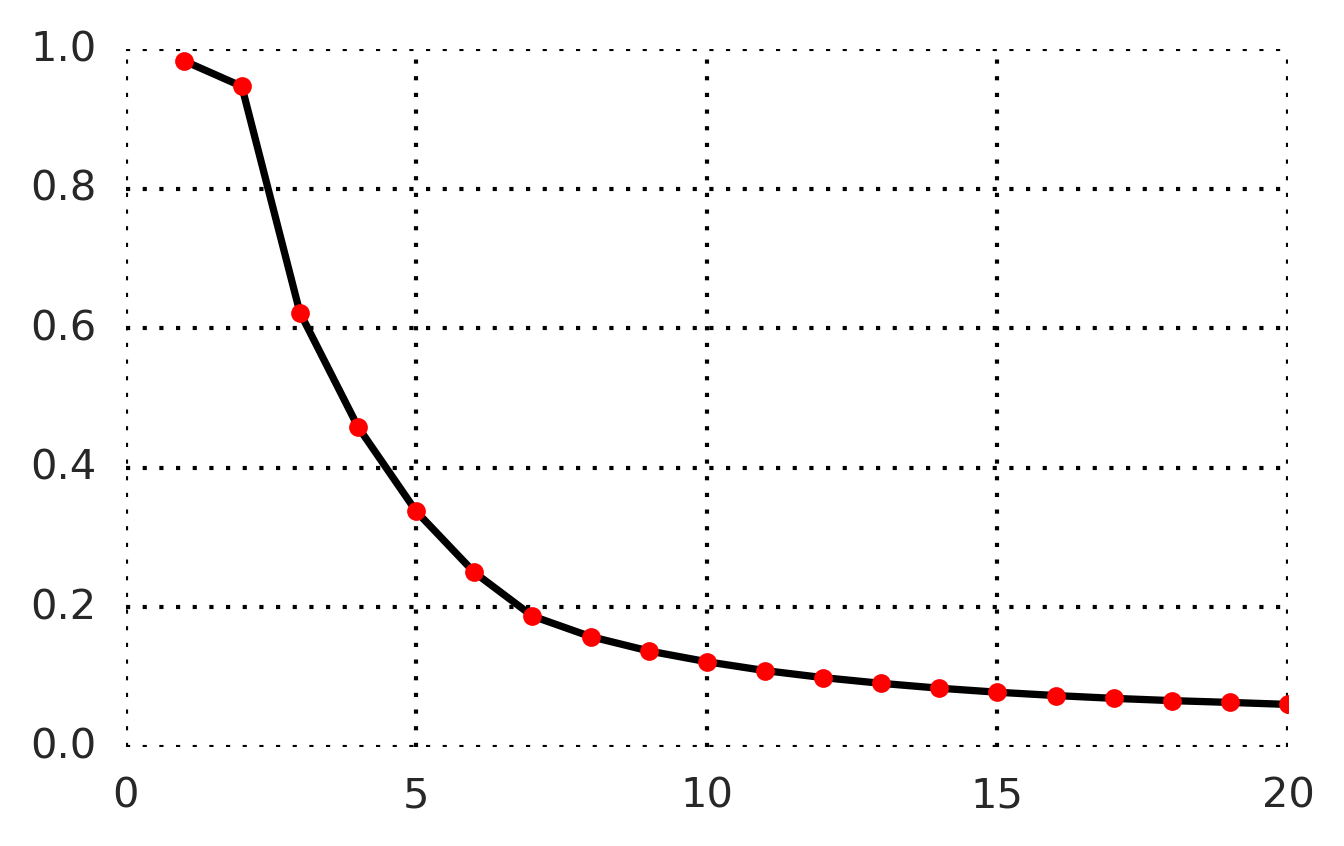}
  \caption{Average accuracy of predicting the positive sample in the contrastive loss for 1 to 20 latent steps in the future of a speech waveform. The model predicts up to 200ms in the future as every step consists of 10ms of audio.}
  \label{pred_acc}
\end{minipage}
\hfill
\end{figure}

\begin{table}[t]
  \hspace*{\fill}
  \begin{minipage}{0.4\textwidth}
      \begin{tabularx}{1.0\textwidth}{l|c}
        \toprule
        \textbf{Method} & \textbf{ACC} \\
        \midrule\midrule
        \textbf{Phone classification} & \\
        Random initialization & 27.6 \\
        MFCC features & 39.7 \\
        CPC & 64.6 \\
        Supervised & 74.6 \\
        \midrule
        \textbf{Speaker classification} & \\
        Random initialization & 1.87 \\
        MFCC features & 17.6 \\
        CPC & 97.4 \\
        Supervised & 98.5 \\    
        \bottomrule
      \end{tabularx}
      \vspace{5pt}
      \caption{LibriSpeech phone and speaker classification results. For phone classification there are 41 possible classes and for speaker classification 251. All models used the same architecture and the same audio input sizes.}
      \label{tab:librispeech}
    \end{minipage}
  \hfill
  \begin{minipage}{0.4\textwidth}
      \centering
      \begin{tabularx}{1.0\textwidth}{l|c}
        \toprule
        \textbf{Method} & \textbf{ACC} \\
        \midrule\midrule
        \textbf{\#steps predicted} & \\
        2 steps & 28.5 \\
        4 steps & 57.6 \\
        8 steps & 63.6 \\
        12 steps & 64.6 \\
        16 steps & 63.8 \\
        \textbf{Negative samples from} & \\
        Mixed speaker & 64.6 \\
        Same speaker & 65.5 \\
        Mixed speaker (excl.) & 57.3 \\
        Same speaker (excl.) & 64.6 \\
        Current sequence only & 65.2 \\
        \bottomrule
      \end{tabularx}
      \vspace{5pt}
      \caption{LibriSpeech phone classification ablation experiments. More details can be found in Section \ref{audio_exp}.
      }
      \label{tab:librispeech_ablation}
    \end{minipage}
    \hspace*{\fill}
    \vspace{-0.5cm}
\end{table}

For audio, we use a 100-hour subset of the publicly available LibriSpeech dataset \cite{panayotov2015librispeech}. Although the dataset does not provide labels other than the raw text, we obtained force-aligned phone sequences with the Kaldi toolkit \cite{povey2011kaldi} and pre-trained models on Librispeech\footnote{www.kaldi-asr.org/downloads/build/6/trunk/egs/librispeech/}. We have made the aligned phone labels and our train/test split available for download on Google Drive\footnote{https://drive.google.com/drive/folders/1BhJ2umKH3whguxMwifaKtSra0TgAbtfb}.
The dataset contains speech from 251 different speakers.

The encoder architecture $g_{enc}$ used in our experiments consists of a strided convolutional neural network that runs directly on the 16KHz PCM audio waveform. We use five convolutional layers with strides [5, 4, 2, 2, 2], filter-sizes [10, 8, 4, 4, 4] and 512 hidden units with ReLU activations. The total downsampling factor of the network is 160 so that there is a feature vector for every 10ms of speech, which is also the rate of the phoneme sequence labels obtained with Kaldi. We then use a GRU RNN \cite{cho2014learning} for the autoregressive part of the model, $g_{ar}$ with 256 dimensional hidden state. The output of the GRU at every timestep is used as the context $c$ from which we predict 12 timesteps in the future using the contrastive loss. We train on sampled audio windows of length 20480. We use the Adam optimizer \cite{kingma2014adam} with a learning rate of 2e-4, and use 8 GPUs each with a minibatch of 8 examples from which the negative samples in the contrastive loss are drawn. The model is trained until convergence, which happens roughly at 300,000 updates.

Figure \ref{pred_acc} shows the accuracy of the model to predict latents in the future, from 1 to 20 timesteps. We report the average number of times the logit for the positive sample is higher than for the negative samples in the probabilistic contrastive loss. This figure also shows that the objective is neither trivial nor impossible, and as expected the prediction task becomes harder as the target is further away.

To understand the representations extracted by CPC, we measure the phone prediction performance with a linear classifier trained on top of these features, which shows how linearly separable the relevant classes are under these features. We extract the outputs of the GRU (256 dimensional), i.e. $c_t$, for the whole dataset after model convergence and train a multi-class linear logistic regression classifier. The results are shown in Table \ref{tab:librispeech} (top). We compare the accuracy with three baselines: representations from a random initialized model (i.e., $g_{enc}$ and $g_{ar}$ are untrained), MFCC features, and a model that is trained end-to-end supervised with the labeled data. These two models have the same architecture as the one used to extract the CPC representations. The fully supervised model serves as an indication for what is achievable with this architecture.
We also found that not all the information encoded is linearly accessible. When we used a single hidden layer instead the accuracy increases from \textbf{64.6} to \textbf{72.5}, which is closer to the accuracy of the fully supervised model.

Table \ref{tab:librispeech_ablation} gives an overview of two ablation studies of CPC for phone classification. In the first set we vary the number of steps the model predicts showing that predicting multiple steps is important for learning useful features. In the second set we compare different strategies for drawing negative sample, all predicting 12 steps (which gave the best result in the first ablation). In the mixed speaker experiment the negative samples contain examples of different speakers (first row), in contrast to same speaker experiment (second row). In the third and fourth experiment we exclude the current sequence to draw negative samples from (so only other examples in the minibatch are present in $X$) and in the last experiment we only draw negative samples within the sequence (thus all samples are from the same speaker).

Beyond phone classification, Table \ref{tab:librispeech} (bottom) shows the accuracy of performing speaker identity (out of 251) with a linear classifier from the same representation (we do not average utterances over time). Interestingly, CPCs capture both speaker identity and speech contents, as demonstrated by the good accuracies attained with a simple linear classifier, which also gets close to the oracle, fully supervised networks.

Additionally, Figure \ref{speaker_tsne} shows a t-SNE visualization \cite{maaten2008visualizing} of how discriminative the embeddings are for speaker voice-characteristics.
It is important to note that the window size (maximum context size for the GRU) has a big impact on the performance, and longer segments would give better results. Our model had a maximum of 20480 timesteps to process, which is slightly longer than a second.

\subsection{Vision}

\begin{figure*}[t!]
  \center
  \includegraphics[width=0.90\textwidth]{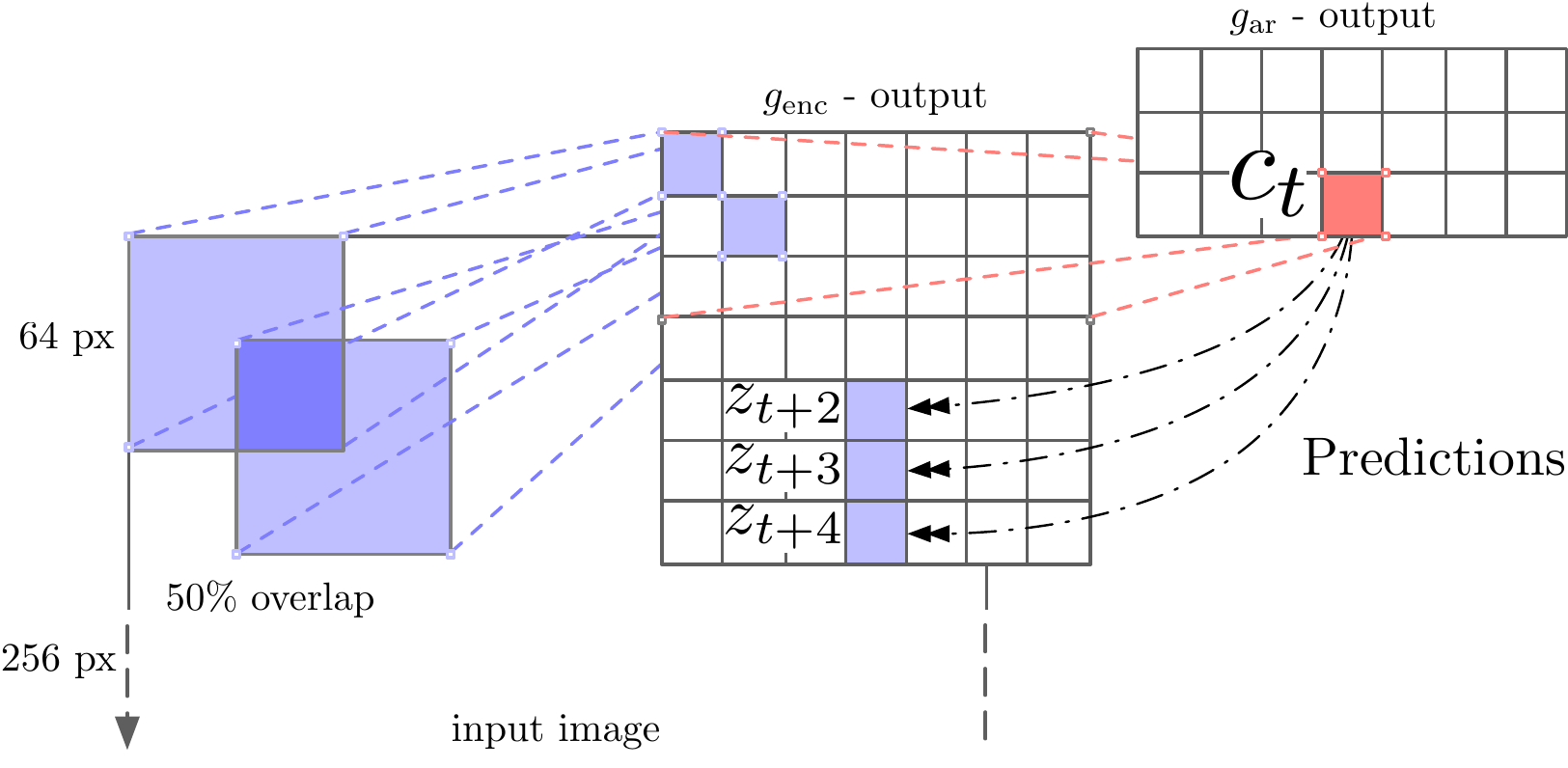}
  \caption{Visualization of Contrastive Predictive Coding for images (2D adaptation of Figure \ref{fig:overview}).}
  \label{fig:overview_img}
\end{figure*}

\begin{figure}[t]
  \centering
  \begin{minipage}{1.0\textwidth}
  \includegraphics[width=1.0\textwidth]{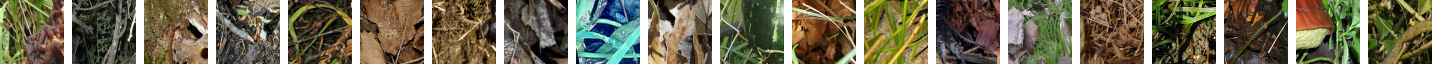}
  \vspace{0.1cm}
  \includegraphics[width=1.0\textwidth]{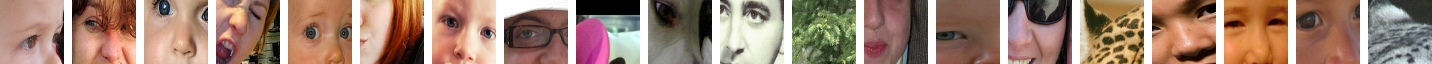}
  \vspace{0.1cm}
  \includegraphics[width=1.0\textwidth]{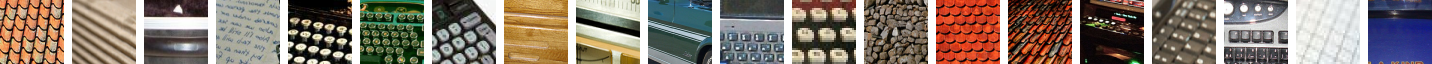}
  \vspace{0.1cm}
  \includegraphics[width=1.0\textwidth]{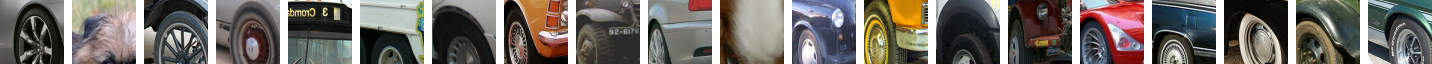}
  \vspace{0.1cm}
  \includegraphics[width=1.0\textwidth]{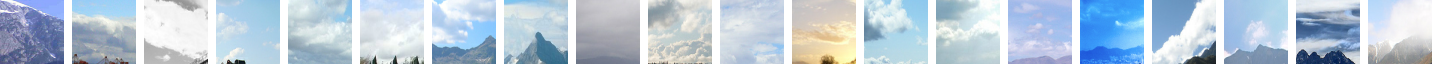}
  \vspace{0.1cm}
  \includegraphics[width=1.0\textwidth]{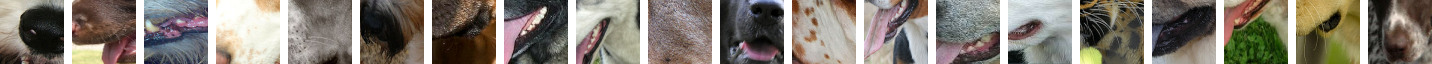}
  \vspace{0.1cm}
  \includegraphics[width=1.0\textwidth]{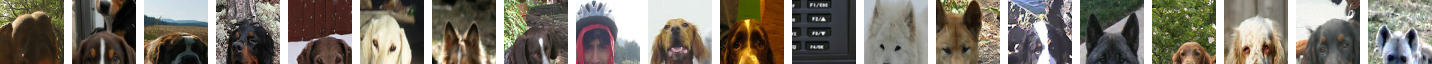}
  \vspace{0.1cm}
  \includegraphics[width=1.0\textwidth]{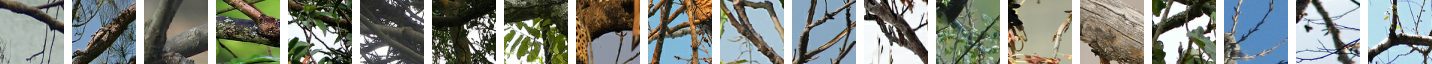}
  \vspace{0.1cm}
  \includegraphics[width=1.0\textwidth]{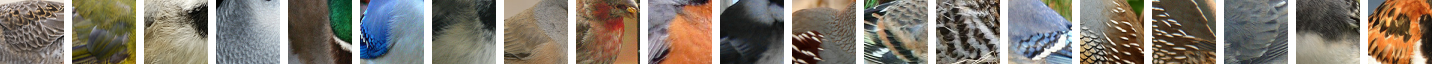}
  \vspace{0.1cm}
  \includegraphics[width=1.0\textwidth]{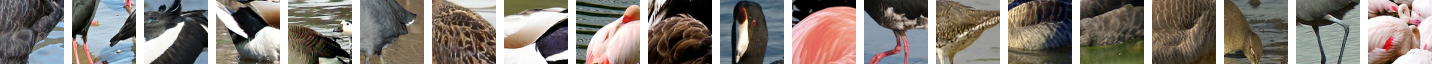}
  \vspace{0.1cm}
  \includegraphics[width=1.0\textwidth]{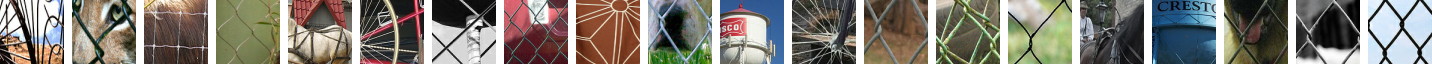}
  \caption{Every row shows image patches that activate a certain neuron in the CPC architecture.}
  \label{neuron_activation}
  \end{minipage}
   \vspace{-0.5cm}
\end{figure}

In our visual representation experiments we use the ILSVRC ImageNet competition dataset \cite{ILSVRC15}. The ImageNet dataset has been used to evaluate unsupervised vision models by many authors \cite{wang2015unsupervised, Doersch_2015_ICCV, donahue2016adversarial, zhang2016colorful, noroozi2016unsupervised, doersch2017multi}. We follow the same setup as \cite{doersch2017multi} and use a ResNet v2 101 architecture \cite{he2016identity} as the image encoder $g_{enc}$ to extract CPC representations (note that this encoder is not pretrained). We did not use Batch-Norm \cite{ioffe2015batch}. After unsupervised training, a linear layer is trained to measure classification accuracy on ImageNet labels.

The training procedure is as follows: from a 256x256 image we extract a 7x7 grid of 64x64 crops with 32 pixels overlap. Simple data augmentation proved helpful on both the 256x256 images and the 64x64 crops. The 256x256 images are randomly cropped from a 300x300 image, horizontally flipped with a probability of 50\% and converted to greyscale. For each of the 64x64 crops we randomly take a 60x60 subcrop and pad them back to a 64x64 image. 

Each crop is then encoded by the ResNet-v2-101 encoder. We use the outputs from the third residual block, and spatially mean-pool to get a single 1024-d vector per 64x64 patch. This results in a 7x7x1024 tensor. Next, we use a PixelCNN-style autoregressive model \cite{aaron2016pixelcnn} (a convolutional row-GRU PixelRNN \cite{aaron2016pixelrnn} gave similar results) to make predictions about the latent activations in following rows top-to-bottom, visualized in Figure \ref{fig:overview_img}. We predict up to five rows from the 7x7 grid, and we apply the contrastive loss for each patch in the row. We used Adam optimizer with a learning rate of 2e-4 and trained on 32 GPUs each with a batch size of 16.

For the linear classifier trained on top of the CPC features we use SGD with a momentum of 0.9, a learning rate schedule of 0.1, 0.01 and 0.001 for 50k, 25k and 10k updates and batch size of 2048 on a single GPU. Note that when training the linear classifier we first spatially mean-pool the 7x7x1024 representation to a single 1024 dimensional vector. This is slightly different from \cite{doersch2017multi} which uses a 3x3x1024 representation without pooling, and thus has more parameters in the supervised linear mapping (which could be advantageous).

Tables \ref{tab:imagenet_top1} and \ref{tab:imagenet} show the top-1 and top-5 classification accuracies compared with the state-of-the-art. Despite being relatively domain agnostic, CPCs improve upon state-of-the-art by 9\% absolute in top-1 accuracy, and 4\% absolute in top-5 accuracy.

\begin{table}[t]
\begin{minipage}{0.47\textwidth}
  \begin{tabularx}{0.99\textwidth}{l|c}
    \toprule
    \textbf{Method} & \textbf{Top-1 ACC} \\
    \midrule\midrule
    \textbf{Using AlexNet conv5} \\
    Video \cite{wang2015unsupervised} & 29.8 \\
    Relative Position \cite{Doersch_2015_ICCV} & 30.4 \\
    BiGan \cite{donahue2016adversarial} & 34.8 \\
    Colorization \cite{zhang2016colorful} & 35.2 \\
    Jigsaw \cite{noroozi2016unsupervised} * & 38.1 \\
    \midrule
    \textbf{Using ResNet-V2} \\
    Motion Segmentation \cite{doersch2017multi} & 27.6 \\
    Exemplar \cite{doersch2017multi} & 31.5 \\
    Relative Position \cite{doersch2017multi} & 36.2 \\ 
    Colorization \cite{doersch2017multi} & 39.6 \\
    \textbf{CPC} & \textbf{48.7} \\
    \bottomrule
  \end{tabularx}
  \vspace{5pt}
  \caption{ImageNet top-1 unsupervised classification results. *Jigsaw is not directly comparable to the other AlexNet results because of architectural differences.}
  \label{tab:imagenet_top1}
\end{minipage}%
\hfill
\begin{minipage}{0.47\textwidth}
  \centering
  \vspace{58pt}
  \begin{tabularx}{0.99\textwidth}{l|c}
    \toprule
    \textbf{Method} & \textbf{Top-5 ACC} \\
    \midrule\midrule
    Motion Segmentation (MS) & 48.3 \\
    Exemplar (Ex) & 53.1 \\
    Relative Position (RP) & 59.2 \\ 
    Colorization (Col) & 62.5 \\
    Combination of & \\
    \quad MS + Ex + RP + Col & 69.3 \\
    \textbf{CPC} & \textbf{73.6} \\
    \bottomrule
  \end{tabularx}
  \vspace{5pt}
  \caption{ImageNet top-5 unsupervised classification results. Previous results with MS, Ex, RP and Col were taken from \cite{doersch2017multi} and are the best reported results on this task.}
  \label{tab:imagenet}
\end{minipage}
\hfill
\end{table}%

\subsection{Natural Language}

\begin{table*}[ht]
  \centering
  \begin{tabularx}{0.73\textwidth}{l|c|c|c|c|c}
    \toprule
    \textbf{Method} & \textbf{MR} & \textbf{CR} & \textbf{Subj} & \textbf{MPQA} & \textbf{TREC}  \\
    \midrule\midrule
    Paragraph-vector \cite{le2014distributed} & 74.8 & 78.1 & 90.5 & 74.2 & 91.8 \\  
    Skip-thought vector \cite{kiros2015skip} & 75.5 & 79.3 & 92.1 & 86.9 & 91.4 \\
    Skip-thought + LN \cite{ba2016layernorm} & 79.5 & 82.6 & 93.4 & 89.0 & -	\\
    \midrule
    CPC & 76.9 & 80.1 & 91.2 &	87.7 & 96.8  \\
    \bottomrule
  \end{tabularx}
  \vspace{5pt}
  \caption{Classification accuracy on five common NLP benchmarks. We follow the same transfer learning setup from Skip-thought vectors \cite{kiros2015skip} and use the BookCorpus dataset as source. \cite{le2014distributed} is an unsupervised approach to learning sentence-level representations. \cite{kiros2015skip} is an alternative unsupervised learning approach. \cite{ba2016layernorm} is the same skip-thought model with layer normalization trained for 1M iterations.}
  \label{tab:textclass}
\end{table*}%

Our natural language experiments follow closely the procedure from \cite{kiros2015skip} which was used for the skip-thought vectors model. We first learn our unsupervised model on the BookCorpus dataset \cite{zhu2015aligning}, and evaluate the capability of our model as a generic feature extractor by using CPC representations for a set of classification tasks.
To cope with words that are not seen during training, we employ vocabulary expansion the same way as \cite{kiros2015skip}, where a linear mapping is constructed between word2vec and the word embeddings learned by the model.

For the classification tasks we used the following datasets: movie review sentiment (MR) \cite{pang2005seeing}, customer product reviews (CR) \cite{hu2004mining}, subjectivity/objectivity \cite{pang2004sentimental}, opinion polarity (MPQA) \cite{wiebe2005annotating} and question-type classification (TREC) \cite{li2002learning}. As in \cite{kiros2015skip} we train a logistic regression classifier and evaluate with 10-fold cross-validation for MR, CR, Subj, MPQA and use the train/test split for TREC. A L2 regularization weight was chosen via cross-validation (therefore nested cross-validation for the first 4 datasets). 

Our model consists of a simple sentence encoder $g_{enc}$ (a 1D-convolution + ReLU + mean-pooling) that embeds a whole sentence into a 2400-dimension vector $z$, followed by a GRU (2400 hidden units) which predicts up to 3 future sentence embeddings with the contrastive loss to form $c$. We used Adam optimizer with a learning rate of 2e-4 trained on 8 GPUs, each with a batch size of 64. We found that more advanced sentence encoders did not significantly improve the results, which may be due to the simplicity of the transfer tasks (e.g., in MPQA most datapoints consists of one or a few words), and the fact that bag-of-words models usually perform well on many NLP tasks \cite{wang2012nlpclassification}.

Results on evaluation tasks are shown in Table \ref{tab:textclass} where we compare our model against other models that have been used using the same datasets. The performance of our method is very similar to the skip-thought vector model, with the advantage that it does not require a powerful LSTM as word-level decoder, therefore much faster to train. Although this is a standard transfer learning benchmark, we found that models that learn better relationships in the childeren books did not necessarily perform better on the target tasks (which are very different: movie reviews etc). We note that better  \cite{zhao2015self,radford2017learning} results have been published on these target datasets, by transfer learning from a different source task.

\subsection{Reinforcement Learning}

\begin{figure*}[t]
  \includegraphics[width=\textwidth]{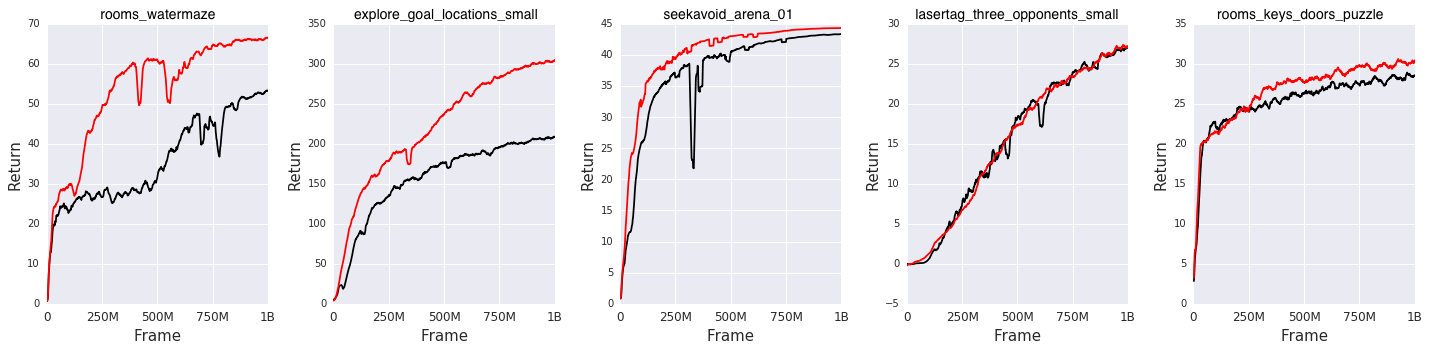}
  \caption{Reinforcement Learning results for 5 DeepMind Lab tasks used in \cite{lasse2018impala}. Black: batched A2C baseline, Red: with auxiliary contrastive loss.}
  \label{fig:dmlab_rl}
  \vspace{-0.3cm}
\end{figure*}

Finally, we evaluate the proposed unsupervised learning approach on five reinforcement learning in 3D environments of DeepMind Lab \cite{beattie2016deepmind}: \mbox{rooms\_watermaze}, \mbox{explore\_goal\_locations\_small}, \mbox{seekavoid\_arena\_01}, \mbox{lasertag\_three\_opponents\_small} and \mbox{rooms\_keys\_doors\_puzzle}.

This setup differs from the previous three. Here, we take the standard batched A2C \cite{mnih2016asynchronous} agent as base model and add CPC as an auxiliary loss. We do not use a replay buffer, so the predictions have to adapt to the changing behavior of the policy. The learned representation encodes a distribution over its future observations.

Following the same approach as \cite{lasse2018impala}, we perform a random search over the entropy regularization weight, the learning-rate and epsilon hyperparameters for RMSProp \cite{hinton2012neural}. The unroll length for the A2C is 100 steps and we predict up to 30 steps in the future to derive the contrastive loss. The baseline agent consists of a convolutional encoder which maps every input frame into a single vector followed by a temporal LSTM. We use the same encoder as in the baseline agent and only add the linear prediction mappings for the contrastive loss, resulting in minimal overhead which also showcases the simplicity of implementing our method on top of an existing architecture that has been designed and tuned for a particular task. We refer to \cite{lasse2018impala} for all other hyperparameter and implementation details. 

Figure \ref{fig:dmlab_rl} shows that for 4 out of the 5 games performance of the agent improves significantly with the contrastive loss after training on 1 billion frames. For \mbox{lasertag\_three\_opponents\_small}, contrastive loss does not help nor hurt. We suspect that this is due to the task design, which does not require memory and thus yields a purely reactive policy.

%% file: conclusion.tex
\section{Conclusion}

In this paper we presented Contrastive Predictive Coding (CPC), a framework for extracting compact latent representations to encode predictions over future observations. CPC combines autoregressive modeling and noise-contrastive estimation with intuitions from predictive coding to learn abstract representations in an unsupervised fashion. 
We tested these representations in a wide variety of domains: audio, images, natural language and reinforcement learning and achieve strong or state-of-the-art performance when used as stand-alone features. The simplicity and low computational requirements to train the model, together with the encouraging results in challenging reinforcement learning domains when used in conjunction with the main loss are exciting developments towards useful unsupervised learning that applies universally to many more data modalities.

\section{Acknowledgements}

We would like to thank Andriy Mnih, Andrew Zisserman, Alex Graves and Carl Doersch for their helpful comments on the paper and Lasse Espeholt for making the A2C baseline available.

%% file: appendix.tex
\appendix

\newpage
\section{Appendix}

\subsection{Estimating the Mutual Information with InfoNCE}

By optimizing InfoNCE, the CPC loss we defined in Equation \ref{loss}, we are maximizing the mutual information between $c_t$ and $z_{t+k}$ (which is bounded by the MI between $c_t$ and $x_{t+k}$). This can be shown as follows.

As already shown in Section \ref{sec:nce}, the optimal value for $f(x_{t+k}, c_{t})$ is given by $\frac{p(x_{t+k}|c_{t})}{p(x_{t+k})}$. Inserting this back in to Equation \ref{loss} and splitting $X$ into the positive example and the negative examples $X_{\text{neg}}$ results in:
\begin{align}
\mathcal{L^{\text{opt}}_{\text{N}}} &= - \mathop{{}\mathbb{E}}_{X} \log\left[  \frac{\frac{p(x_{t+k}|c_{t})}{p(x_{t+k})}}{\frac{p(x_{t+k}|c_{t})}{p(x_{t+k})} + \sum_{x_j \in X_{\text{neg}}} \frac{p(x_{j}|c_{t})}{p(x_{j})}}\right] \\
&= \mathop{{}\mathbb{E}}_{X}\log\left[1 + \frac{p(x_{t+k})}{p(x_{t+k}|c_{t})} \sum_{x_j \in X_{\text{neg}}} \frac{p(x_{j}|c_{t})}{p(x_{j})}\right] \\
&\approx \mathop{{}\mathbb{E}}_{X}\log\left[1 + \frac{p(x_{t+k})}{p(x_{t+k}|c_{t})} (N-1) \mathop{{}\mathbb{E}}_{x_j} \frac{p(x_{j}|c_{t})}{p(x_{j})}\right] \label{mi_approx} \\
&= \mathop{{}\mathbb{E}}_{X}\log\left[1 + \frac{p(x_{t+k})}{p(x_{t+k}|c_{t})} (N-1)\right] \\
&\geq \mathop{{}\mathbb{E}}_{X}\log\left[\frac{p(x_{t+k})}{p(x_{t+k}|c_{t})} N\right] \label{mi_ineq} \\
&= - I(x_{t+k}, c_t) + \log(N),
\end{align}
Therefore, $I(x_{t+k}, c_t)\geq \log(N) - \mathcal{L^{\text{opt}}_{\text{N}}}$. This trivially also holds for other $f$ that obtain a worse (higher) $\mathcal{L}_{\text{N}}$. Equation \ref{mi_approx} quickly becomes more accurate as $N$ increases. At the same time $\log(N) - \mathcal{L_{\text{N}}}$ also increases, so it's useful to use large values of $N$. 

InfoNCE is also related to MINE \cite{belghazi2018mine}. Without loss of generality let's write $f(x, c) = e^{F(x, c)}$, then
\begin{align}
\mathop{{}\mathbb{E}}_{X}\left[ \log \frac{f(x, c)}{\sum_{x_j \in X} f(x_j, c)}\right] 
&= \mathop{{}\mathbb{E}}_{(x, c)} \Big[F(x, c)\Big] - \mathop{{}\mathbb{E}}_{(x, c)}\Big[\log\sum_{x_j \in X} e^{F(x_j, c)}\Big] \\
&= \mathop{{}\mathbb{E}}_{(x, c)} \Big[F(x, c)\Big] - \mathop{{}\mathbb{E}}_{(x, c)}\Big[\log \Big(e^{F(x, c)} + \sum_{x_j \in X_{\text{neg}}} e^{F(x_j, c)}\Big)\Big] \\
&\leq \mathop{{}\mathbb{E}}_{(x, c)} \Big[F(x, c)\Big] - \mathop{{}\mathbb{E}}_{c}\Big[\log \sum_{x_j \in X_{\text{neg}}} e^{F(x_j, c)}\Big] \\
&= \mathop{{}\mathbb{E}}_{(x, c)} \Big[F(x, c)\Big] - \mathop{{}\mathbb{E}}_{c}\Big[\log \frac{1}{N-1}\sum_{x_j \in X_{\text{neg}}} e^{F(x_j, c)} + \log(N-1)\Big]
\end{align}
is equivalent to the MINE estimator (up to a constant). So we maximize a lower bound on this estimator. We found that using MINE directly gave identical performance when the task was non-trivial, but became very unstable if the target was easy to predict from the context (e.g., when predicting a single step in the future and the target overlaps with the context).